\documentclass[twocolumn]{article}
\usepackage{graphicx} 
\usepackage{booktabs}
\usepackage{natbib}
\usepackage{multicol}
\usepackage{url}
\usepackage[a4paper, total={6in, 9in}]{geometry}
\usepackage[width=0.9\columnwidth]{caption}
\usepackage{makecell}
\usepackage{tablefootnote}
\usepackage{tabularray}
\usepackage[table]{xcolor}

\title{\Large{Towards the first UD Treebank of Spoken Italian: the \textit{KIParla forest}}}
\author{\small{Ludovica Pannitto, Experimental Laboratory - University of Bologna - Italy}}
\date{\small{April 2024}}

\begin{document}

\maketitle

The present project endeavors to enrich the linguistic resources available for Italian by constructing a treebank for the KIParla corpus \citep{maurikiparla,ballare2020creazione}, an existing and well known resource for spoken Italian.
At the moment of writing, in fact, none of the available italian UD treebanks is explicitly addressing spoken varieties. In particular, the Venice Italian Treebanks, VIT \citep{delmonte2007vit}, contains approximately 60k tokens of spoken dialogues from different Italian varieties: they were however collected already 20 years ago now and as the result of a Map Task, therefore not in ecological, conversational situation. Other treebanks such as ParTUT \citep{sanguinetti2015parttut} contain portions of controlled transcribed speech, such as transcribed Ted Talks \citep{cettolo2012wit3}. Transcripts of formal speeches also appear in the ParlaMINT treebank \citep{agnoloni2022making}.
Differently from the aforementioned resources, the genesis of the KIParla corpus stems from the imperative to capture the linguistic diversity inherent in spoken language, representing variation with respect to the properties of the interaction and the properties of the speakers involved. Geographical differentiation is represented in the corpus both by the influence and occasional use of local dialects and by regional traits, perceptible even in the speech of educated individuals. 
More generally, spoken data are rare in the broader context of Universal Dependency treebanks \citep{Dobrovoljc2022} but offer unique insights into linguistic phenomena and support various research fields, ranging from typology to NLP. A greater availability of spoken treebanks would open the path to large-scale studies on phenomena such as conversational patterns, discourse markers, and syntactic variation, which are hard to scale above the lexical level with available resources. From the NLP perspective, currently available pipelines often drastically drop their accuracy rate when running on spoken language varieties and no spoken resource is currently available to train accurate annotation pipelines tailored to speech data.

The KIParla corpus, 
initially a collaborative effort between the University of Bologna and the University of Turin, encompasses a diverse range of Italian spoken varieties, manually transcribed following Jefferson guidelines\footnote{\url{https://benjamins.com/catalog/pbns.125.02jef}} and aligned with audio files.
The main novelty of KIParla in the landscape of spoken data collection is that it is structured in an incremental and modular fashion, which allows for the addition of new corpus modules over time, focusing on different dimensions of linguistic variation and geographical areas. Thanks to thoughtful guidelines on formats, transcriptions and metadata, the integration of new modules into the overall resource is easily achieved. The entire corpus is freely available for consultation\footnote{\url{http://kiparla.it}}, and an orthographic transcription is already provided upon request. Each module within the overarching KIParla project is balanced to ensure its coherence and self-sufficiency as a linguistic resource (Table \ref{tab:moduli}).

\begin{table}[h!]
\centering
\resizebox{\columnwidth}{!}{%

\begin{tabular}{@{}ll@{}}
\toprule
\textbf{Module} & 
\textbf{Description}  \\ \midrule

KIP\tablefootnote{\cite{goria2018corpus}} & 
\makecell[l]{data from university lectures, examinations, student consultations, \\semi-structured interviews, and spontaneous speech. \\Duration: 70 hours, 661,175 tokens. Status: Available. }\\ \midrule
ParlaTO\tablefootnote{\cite{massimo2020parlato}} & 
\makecell[l]{individual interviews and group discussions. \\Duration: 50 hours. Status: Available. }\\ \midrule
ParlaBO & 
\makecell[l]{semi-structured interviews. \\Duration: 50 hours. Status: Close to publication.}\\ \midrule
ParlaBZ & 
\makecell[l]{semi-structured interviews and dinner table conversations.\\ Duration: Approximately 5 hours. Status: Close to publication. }\\ \midrule
KIPasti & 
\makecell[l]{spontaneous speech during family or friendly gatherings.\\ Duration: 40 hours, 487,607 tokens. Status: Available }\\ \midrule
Stra-ParlaBO & 
\makecell[l]{semi-structured interviews and spontaneous speech from speakers\\ with international migration backgrounds.\\ Duration: Approximately 50 hours. Status: Under construction. }\\ \midrule
Stra-ParlaTO & 
\makecell[l]{semi-structured interviews and spontaneous speech from speakers\\ with international migration backgrounds.\\ Duration: Approximately 50 hours. Status: Under construction.} \\ \midrule
ParlaMI & 
\makecell[l]{semi-structured interviews and dinner table conversations.\\ Duration: Approximately 60 hours. Status: Planned.} \\ \midrule
ParlaNA & 
\makecell[l]{various communicative contexts. Status: Planned.} \\ \bottomrule

\end{tabular}
}
\caption{KIParla Project Modules}
\label{tab:moduli}
\end{table}

Preliminary linguistic annotation efforts on the KIParla corpus were initiated during the EVALITA\footnote{\url{https://www.evalita.it/}} evaluation campaign in 2020. The KIPoS task \citep{Bosco2020}, focusing on Part-of-Speech tagging of KIParla data, received attention within this campaign. The dataset provided for the task comprised approximately 200,000 tokens, sourced from the KIP module, with around 30,000 tokens undergoing manual review and correction after automatic processing through UDPipe \citep{straka-2018-udpipe}.

Participating teams obtained satisfactory results on the task, highlighting the challenges in morphologically and syntactically tagging speech, especially in unconstrained speech contexts.
Those will especially be the focus of my participation in the training school and could be addressed during the brainstorming hackathon, namely: the identification of mode-specific phenomena, tracing the greater variety of non-standard and creative alternatives to be assigned back to the same linguistic phenomenon (e.g., forms to be assigned to the same lemma), and dealing with different types of interaction and registers.
The newly collected modules, more specifically the KIPasti one and those targeting international migration, are bound to introduce \textbf{further} challenges, especially concerning code-switching phenomena both towards local dialects and towards other languages present in the speakers' linguistic background. In the KIPoS task, for instance, two Part-of-Speech labels were introduced to tag tokens coming from \textit{Italo-Romance dialects} and from \textit{Languages other than Italian} (namely, \textsc{DIA} and \textsc{LIA} respectively). Revisions may be necessary to accommodate structural code-switching phenomena, aligning with approaches in other multilingual UD treebanks \citep{ccetinouglu2019challenges,Braggaar2021}.
As pointed out in \cite{Dobrovoljc2022}, many other aspects need specific attention in the case of spoken treebanks: in the case of KIParla, some of them have already been addressed in the creation of the resource. For instance, as far as \textbf{orthography and punctuation} are concerned, the corpus is already provided with a standard orthography transcription, but no sentence-initial capitalization is performed and standard punctuation is inserted in cases of questions and exclamations.
\textbf{Segmentation} into sentence-like units is notoriously a controversial topic when it comes to speech. In the KIPoS task the original intonation-based segmentation was taken to define sentence boundaries: the Rhapsodie treebank \citep{kahane2021annotation} considers instead illocutory units (i.e., a speech segment that corresponds to a single speech act) as basic units for segmentation, thus differentiating between microsyntactic (government) units and macrosyntactic maximal units \citep{pietrandrea2014notion}. This approach seem to present numerous advantages, namely the fact that it proceeds bottom-up building dependency relations among tokens and that it can be more easily operationalized in a series of tests to guide annotators.
A connected aspect is that of \textbf{speakers' overlap}, which is also taken track of in the Rhapsodie approach. Table \ref{tab:features} summarizes the different approaches taken in available UD spoken treebanks with respect to the mentioned aspects: the last column reports features that a treebank based on KIParla could provide.

\begin{table}[htbp]
    \centering

    \resizebox{\columnwidth}{!}{%
\begin{tabular}{@{}l|llllllllllll|l@{}}
 &
  \rotatebox{90}{Beja} &
  \rotatebox{90}{Cantonese} &
  \rotatebox{90}{Chinese} &
  \rotatebox{90}{Chukchi} &
  \rotatebox{90}{ParisStories} &
  \rotatebox{90}{Rhapsodie} &
  \rotatebox{90}{Frisian-Dutch} &
  \rotatebox{90}{Komi-Zyrian} &
  \rotatebox{90}{Naija} &
  \rotatebox{90}{Norwegian} &
  \rotatebox{90}{Slovenian} &
  \rotatebox{90}{\makecell[l]{Turkish-\\German}} &
  \rotatebox{90}{\makecell[l]{\textbf{KIParla}\\ \textbf{forest}}}\\ \midrule
Sound file ID        & yes & no  & no  & yes & yes & no  & no  & no  & yes & no  & no  & no & yes \\
Text-sound alignment & yes & no  & no  & yes & no  & no  & no  & no  & yes & no  & no  & no & yes \\
Speaker ID           & no  & no  & no  & no  & yes & yes & yes & no  & yes & yes & no  & no & yes \\
Language variety     & no  & no  & no  & no  & no  & no  & yes & yes & no  & yes & no  & yes & yes \\ \hline
Standard ortography  & no  & no  & yes & yes & yes & yes & yes & no  & no  & yes & yes & yes & yes \\
Capitalization       & no  & no  & no  & yes & no  & no  & no  & yes & no  & no  & no  & yes & no \\
Pronunciation        & yes & no  & no  & yes & no  & no  & no  & no  & no  & no  & yes & no  & no \\ \hline
Speaker overlap      & no  & no  & no  & no  & no  & yes & no  & no  & no  & no  & yes & no  & yes \\
Final punctuation    & yes & yes & yes & yes & yes & yes & no  & yes & yes & yes & no  & yes & no \\
Other punctuation    & yes & yes & yes & no  & yes & yes & no  & yes & yes & yes & no  & yes & no \\ \hline
Incomplete words     & no  & no  & no  & yes & yes & yes & no  & no  & yes & yes & yes & yes & yes \\
Fillers              & no  & no  & no  & no  & yes & yes & yes & no  & yes & yes & yes & yes & yes \\
Silent pauses        & yes & no  & no  & no  & no  & no  & no  & no  & yes & yes & yes & no  & yes \\
Incidents            & no  & no  & no  & no  & no  & no  & no  & no  & no  & no  & yes & no & yes \\ \bottomrule
\end{tabular}%
}
\caption{Features to be found in spoken UD treebanks, from \cite{Dobrovoljc2022}}
\label{tab:features}
\end{table}

The proposal is therefore to build on the already existing portion of annotated data for the KIPoS task, enlarging it with 30,000 tokens coming from the new conversational contexts from the KIPasti module, therefore planning a first release consisting of approximately 60,000 tokens. Participation in the training school will be crucial for the project, because it will allow us to focus on the challenges just highlighted, discussing the possible solutions with expert scholars. 
Given its size and the variety of conversational contexts recorded in the KIParla, the resulting treebank would represent an invaluable resource to investigate syntax-level phenomena in relation to regional variation, standard vs. substandard and formal vs. informal varieties, sociolectal variation and multilingualism, providing us with the first large scale picture of the linguistic richness and complexity of variation in spoken Italian and its usage across different contexts and communities.

\bibliography{references}

\end{document}